\documentclass[10pt,twocolumn,letterpaper,table]{article}

\usepackage{cvpr}      %

\usepackage{graphicx}
\usepackage{amsmath}
\usepackage{amssymb}
\usepackage{booktabs}
\usepackage{multirow}
\usepackage{multicol}
\usepackage{tikz}
\usepackage{pgfplots}
\pgfplotsset{compat=1.18} 
\usepackage{xcolor}
\usepackage{soul}
\usepackage{algorithm}
\usepackage{algpseudocode}
\usepackage{subfigure}
\usepackage{overpic}
\usepackage{pifont}%
\usepackage[accsupp]{axessibility}
\newcommand{\cmark}{\ding{51}}%
\newcommand{\xmark}{\ding{55}}%
\definecolor{cvprblue}{rgb}{0.21,0.49,0.74}
\usepackage[pagebackref,breaklinks,colorlinks,citecolor=cvprblue]{hyperref}

\usepackage[capitalize]{cleveref}
\crefname{section}{Sec.}{Secs.}
\Crefname{section}{Section}{Sections}
\Crefname{table}{Table}{Tables}
\crefname{table}{Tab.}{Tabs.}
\definecolor{myblue}{RGB}{24, 131, 192}
\definecolor{myorange}{RGB}{255, 136, 0}
\definecolor{mygreen}{RGB}{37, 171, 20}
\definecolor{myred}{RGB}{222,35,38}
\definecolor{mybrown}{RGB}{160, 113, 20}

\def\cvprPaperID{1238} %
\def\confName{CVPR}
\def\confYear{2024}

\begin{document}

\title{Coherent Temporal Synthesis for Incremental Action Segmentation}

\author{Guodong Ding, Hans Golong and Angela Yao\\
National University of Singapore\\
{\tt\small \{dinggd, hgolong, ayao\}@comp.nus.edu.sg}
}
\maketitle

\begin{abstract}
Data replay is a successful incremental learning technique for images. It prevents catastrophic forgetting by keeping a reservoir of previous data, original or synthesized, to ensure the model retains past knowledge while adapting to novel concepts. However, its application in the video domain is rudimentary, as it simply stores frame exemplars for action recognition. 
This paper presents the first exploration of video data replay techniques for incremental action segmentation, focusing on action temporal modeling. We propose a Temporally Coherent Action (TCA) model, which represents actions using a generative model instead of storing individual frames. The integration of a conditioning variable that captures temporal coherence allows our model to understand the evolution of action features over time. Therefore, action segments generated by TCA for replay are diverse and temporally coherent. In a 10-task incremental setup on the Breakfast dataset, our approach achieves significant increases in accuracy for up to 22\% compared to the baselines. 
\end{abstract}

\section{Introduction}\label{sec:intro}

Ensuring that intelligent systems can continually adapt and accumulate knowledge in our rapidly evolving world is essential. This concept is embodied in incremental learning~\cite{de2021continual}. A key challenge is to acquire new knowledge gradually without 
catastrophic forgetting~\cite{french1999catastrophic} 
of previously learned information. There have been numerous efforts to tackle the problem in the machine learning community, including data replay~\cite{rebuffi2017icarl}, regularization techniques~\cite{kirkpatrick2017overcoming}, and knowledge distillation~\cite{li2017learning}.

Data replay is an effective and commonly used technique in image incremental learning~\cite{xiang2019incremental,shin2017continual,aljundi2019gradient,bang2021rainbow}. Data replay mitigates catastrophic forgetting by re-exposing the model to previously encountered data samples. Previous works either retain a subset of the original training samples~\cite{hou2019learning,lopez2017gradient,rebuffi2017icarl} or learn a generative model on previous training data to later generate surrogate training samples~\cite{hayes2020remind,lesort2019generative,shin2017continual}. 
A recent shift of incremental learning from static images to dynamic videos has brought about the application of data replay into the video domain~\cite{villa2022vclimb,alssum2023just,park2021class}. However, this transition has been limited to direct implementations, with less emphasis on videos' unique temporal properties.

\textit{Action recognition}~\cite{kong2022human}, as the hallmark task of video understanding, was the initial focus for incremental learning for videos~\cite{villa2022vclimb,alssum2023just,park2021class,pei2022learning}. 
Many works create replay data by storing frames for each video clip. Storing frame-exemplars for videos requires a balance between \textit{dense temporal sampling} and \textit{memory diversity}~\cite{villa2022vclimb,alssum2023just}. It is a trade-off between storing fewer complete videos or more incomplete ones within a given memory budget. 
Because existing recognition models accept sparse frames as input, like 8 or 16 frames for TSN~\cite{wang2016temporal} and TSM~\cite{lin2019tsm}, keeping an exemplar set for each action clip does not overly strain the memory allocation. Also, there is a strong static bias towards scene context~\cite{li2018resound} in standard action recognition datasets like Kinetics~\cite{carreira2017quo} and UCF101~\cite{soomro2012ucf101}. 
It is, therefore, possible to achieve comparable performance storing either entire segments, a few frames~\cite{villa2022vclimb}, or even a single frame~\cite{alssum2023just}, all without the need for explicit temporal modeling. 
As a result, most techniques save exemplar frames and mirror the practices from the image domain while giving less attention to developing video-specific data replay techniques.

We advocate exploring data replay strategies in incremental video understanding with \textit{temporal action segmentation} (TAS)~\cite{ding2023temporal}. 
TAS is akin to semantic image segmentation~\cite{minaee2021image} but from a one-dimensional temporal basis. 
Instead of 2D pixel-wise semantic labels, TAS assigns frame-level action labels. For instance, given a procedural video of ``\textit{making coffee}'', a TAS model classifies each frame as a step such as `\textit{take cup}', `\textit{pour coffee}', `\textit{pour milk}', \etc. In TAS, a procedural task is usually called an ``\textit{activity}'' and the steps `\textit{actions}'. 
The necessity of temporal modeling in TAS makes it more suitable for investigating video-specific replay strategies. First, static bias is less prominent in TAS since different actions within a sequence often share a common background, emphasising the importance of temporal modeling for video data replay.  
Second, TAS operates with full-resolution frame-by-frame inputs and produces outputs at the same level of temporal granularity. The extended temporal span requires more capacity in the replay memory compared to shorter, trimmed action clips. Storing frame exemplars in proportion to the video's duration, even downsampled, imposes a significant memory burden. Using sparse frames to represent actions also results in a loss of temporal coherence in the natural progression of actions. %

We present a novel temporally coherent action model for video data replay in incremental action segmentation. Contrary to the conventional frame-exemplar storing approaches, we use a generative model to represent actions. This is motivated by the model's capacity to learn efficient data representations with a fixed model size while producing diverse outputs of arbitrary lengths. 
The replay data generation is top-down. First, a replay video is defined by its sequential structure, including action sequences and segment durations. Subsequently, the generative model generates features for each action segment. Finally, these generated segments are concatenated to form a complete replay video. 
Unlike generative approaches for image tasks~\cite{hayes2020remind,lesort2019generative,shin2017continual}, our model incorporates a conditioning variable to account for the unique temporal coherence of videos. The coherence variable, defined as the relative progression within an action, assists the model in capturing the evolution of action features over time.

\noindent \textbf{Contributions.} Our contributions are summarized as follows: 
\textbf{(1)} To the best of our knowledge, we are the first to introduce the incremental action segmentation task, working with procedural videos. This is a natural fit for the development of intelligent assistants which learn complex tasks and activities in the real world in an incremental manner. 
\textbf{(2)} We propose to model actions with generative models and use the models to generate diverse replay data. The proposed generative data replay bypasses the trade-off problem in frame-exemplar storing approaches. 
\textbf{(3)} We introduce a temporal coherence variable to help the generative model learn action feature evolution and produce temporally coherent action segments in replay video generation. 
\textbf{(4)} Experiments on two procedural benchmarks show that our approach can effectively mitigate catastrophic forgetting for incrementally learned action segmentation models.

\section{Related Works}\label{sec:related}
 
\noindent \textbf{Incremental Learning.} 
Incremental learning involves algorithms adapting to new data without forgetting prior knowledge. Popular approaches in the image domain include data replay~\cite{rebuffi2017icarl}, regularization techniques~\cite{kirkpatrick2017overcoming}, and knowledge distillation~\cite{li2017learning}. 
Data replay methods in the image domain are categorized into direct replay, using the original exemplar set for rehearsal~\cite{hou2019learning,lopez2017gradient,rebuffi2017icarl}, and generative replay, which models sample distribution to generate instances~\cite{hayes2020remind,lesort2019generative,shin2017continual}. Limited attention has been given to the video domain~\cite{villa2022vclimb,alssum2023just}, and most works rely on direct replay of frame-wise exemplars while neglecting temporal aspects. This work proposes leveraging generative models, focusing on maintaining temporal coherence for video data replay.

\noindent \textbf{Temporal Action Segmentation.} 
There have been many existing approaches proposed for the temporal action segmentation task. Fully supervised approaches rely on the dense annotation of the video frames~\cite{farha2019ms,yi2021asformer}. In a semi-supervised setting~\cite{singhania2022iterative,ding2022leveraging}, only a subset of videos requires dense labels, while the remaining videos are unlabeled. Weaker forms of supervision include action transcripts~\cite{kuehne2017weakly}, action sets~\cite{richard2018action,li2020set,fayyaz2020sct}, timestamps~\cite{li2021temporal,rahaman2022generalized} and activity labels~\cite{ding2022temporal}. Some cases~\cite{sarfraz2021temporally,sener2018unsupervised,kukleva2019unsupervised} work without any action labels in an unsupervised setup. Despite the ongoing efforts in TAS with diverse forms of supervision, incremental learning has not been explored. Our work is the first to study incremental action segmentation, emphasizing video replay techniques. %

\section{Preliminaries}\label{sec:pre}
\noindent \textbf{Temporal Action Segmentation} (TAS) divides untrimmed video sequences into temporal segments and associates each segment with a predefined action label~\cite{ding2023temporal}. Given a video $x^{1:T} = \{x^1,..., x^T\}$ of $T$ frames long, a model $\mathcal{M}$ segments $x$ into $N$ contiguous and non-overlapping actions: 
\begin{equation}\label{eq:segment}
    \mathbf{s}_{1:N} = (s_1,s_2,...,s_N), \quad \text{where }   
    s_n = (a_n, t_n, \ell_n),
\end{equation}
\begin{equation}
   \text{s.t.} \qquad t_{n+1} = t_n+\ell_n, \nonumber
\end{equation}
$s_n$ denotes a temporal segment in the video of length of $\ell_n$, with action class label $a_n \in \mathcal{A}$ from $A$ predefined categories. $t_n$ denotes the starting timestamp of segment $s_n$ and adheres to a precise temporal sequence from the preceding segment. In practice, most existing works~\cite{lea2017temporal,farha2019ms,singhania2023c2f,yi2021asformer} design $\mathcal{M}$ to classify actions on a per-frame basis, \ie,
\begin{equation}\label{eq:framewise}
    y^{1:T}= (y^1,y^2,...,y^T),
\end{equation}
where $y^t \in \mathcal{A}$ is the frame-wise action label at time $t$.  

Common architectures for the segmentation model $\mathcal{M}$ include convolution-based MSTCN~\cite{farha2019ms} and transformer-based ASFormer~\cite{yi2021asformer}. Due to the memory constraints, $x^t$ is typically provided as pre-computed features such as I3D~\cite{carreira2017quo} rather than raw RGB inputs. The segmentation model $\mathcal{M}$ is trained with a frame-wise cross-entropy loss:
\begin{equation}\label{eq:cls}
    \mathcal{L}_{\text{cls}}(x,y) =  \frac{1}{T} \sum_t -\log(p^t(y^t)), 
\end{equation}
where $p^t \in \mathbb{R}^{A}$ is the estimated action probabilty for the frame $x^t$. In addition, a smoothing loss is imposed to ensure smooth transitions between consecutive frames:
\begin{equation}\label{eq:tmse}
    \mathcal{L}_{\text{sm}}(x,y) = \frac{1}{TA}\sum_{t,a}\tilde{\Delta}_{t,a}^2, \;\;
    \tilde{\Delta}_{t,a} = \begin{cases}
    \Delta_{t,a}\kern-0.8em &:\! \Delta_{t,a} \le \tau\\
    \tau \kern-0.8em&:\! \text{otherwise}
    \end{cases},
\end{equation}
\begin{equation}
    \Delta_{t,a} = \left|\log p^t(a) - \log p^{t-1}(a)\right|.\nonumber
\end{equation}
$\tau$ is conventionally set to 4, as per~\cite{farha2019ms}. The full training loss is written as the combination of the above two, balanced by hyperparameter $\lambda$:
\begin{equation}\label{eq:tas}
    \mathcal{L}_{\text{tas}} = \mathcal{L}_{\text{cls}} + \lambda\cdot \mathcal{L}_{\text{sm}}.
\end{equation}

\noindent \textbf{Class Incremental Learning} (CIL) introduces new classes to model $\mathcal{M}$ over time~\cite{rebuffi2017icarl}. Consider a series of $B$ tasks, each task $\mathcal{T}_b$ represents a distinct incremental step, comprising a set of $n_b$ training instances denoted as $\{(x_i, y_i)\}_{i=1}^{n_b}$. Here, $x_i$ represents an instance associated with class $y_i\in\mathbf{Y}_b$, where $\mathbf{Y}_b$ is the class set for task $b$. Importantly, access to the training data in $\mathcal{T}_b$ is restricted to training task $b$ only, and there is no longer access to data from tasks prior to $b$. The developed model must acquire knowledge from the current task while preserving knowledge from the past tasks. The performance of an incremental learning model is evaluated over all seen classes, \ie, $\mathcal{Y}_B=\mathbf{Y}_1\cup... \mathbf{Y}_B$. 
Standard CIL assumes non-overlapping classes in different tasks ($\mathbf{Y}_b\cap\mathbf{Y}_b'=\O$ for $b\neq b'$), but when class overlap occurs, the task referred to as blurry class-incremental learning (Blurry CIL)~\cite{koh2021online,bang2021rainbow,bang2022online}. 

Data replay is a commonly adopted approach for incremental learning by revisiting former exemplars.  The exemplar set, also known as the replay buffer, is an extra collection of instances from the previous tasks $\hat{\mathcal{T}}_{1:b} = \{(\mathbf{x}_{b'}, \mathbf{y}_{b'})\}_{b'=1}^{b-1}, y_{b'} \in \mathcal{Y}_{b-1}$. 
The exemplars can either be specified training samples~\cite{rebuffi2017icarl}, constructed prototypes~\cite{iscen2020memory} or through  generation~\cite{shin2017continual,xiang2019incremental}. The model can then utilize $\mathcal{T}_b \cup \hat{\mathcal{T}}_{1:b}$ for update while attaining previous knowledge.

\section{Incremental Temporal Action Segmentation} \label{sec:itas}
Incremental temporal action segmentation (iTAS) continuously updates an action segmentation model when encountering new action classes over time. Unlike existing CIL task that assumes no interclass dependencies, iTAS works with procedural videos where the same activity often share a common set of actions, making it more intuitive to treat each procedural activity as an incremental task $b$ and the actions within each activity as class labels $\mathbf{Y}_b$. The segmentation model $\mathcal{M}$ learns to segment videos from various incremental activities.
This work introduces a novel generative data replay approach for iTAS featuring a temporally coherent action model.

\subsection{Temporally Coherent Action Modeling}
Generative models, such as Variational Autoencoders (VAE)~\cite{kingma2013auto} and Generative Adversarial Networks~\cite{goodfellow2014generative} are powerful tools for learning efficient representations of data with a fixed model size. Additionally, they are easy to extend to a conditional form to generate conditioned outputs.
These properties make them a good fit for the purpose of generating diverse action segments for data replay. 

In this work, we employ a conditional VAE to model the actions. Other generative models are also feasible, but conditional VAEs offer a good tradeoff between model size, expressiveness, and data efficiency for learning. In a conditional VAE model, an encoder maps an input frame feature $x$ and the conditioning action class label $a$ to a probability distribution $q_{\phi}(z|x, a)$ over the latent space, where $z$ is the latent variable. The decoder maps a sample from latent space $z$ and the conditioning information back to obtain a reconstruction of the input $\hat{x}= p_{\theta}(x|z, a)$. $\phi,\theta$ are learnable network parameters of the decoder and encoder, respectively. The overall loss function is a combination of the reconstruction term and the KL divergence regularization term, written as:
\begin{equation}\label{eq:cvae}
    \mathcal{L}_{\text{cVAE}} = \underbrace{\mathbb{E}_z \log p_{\theta}(x|z,a)}_{\text{reconstruction}} - \underbrace{\text{D}_{\text{KL}}(q_{\phi}(z|x,a)||p(z))}_{\text{regularization}}.
\end{equation}
The second term regularizes the latent space by a prior $p(z)$ on the approximate posterior distribution; typically, the prior is a Gaussian distribution. 

\begin{figure}
    \centering
    \begin{overpic}[width=\linewidth]{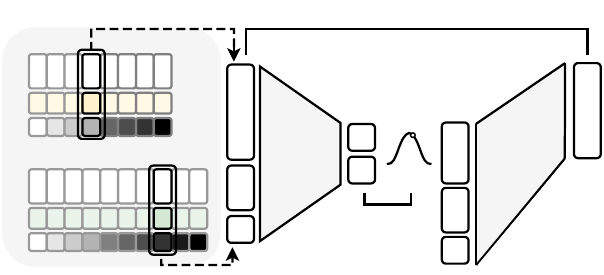}

    \put(5.3,37.4){\footnotesize $1$}
    \put(25.5,37.4){\footnotesize $\ell_1$}
    \put(0.7,33){\footnotesize $\mathbf{x}_1$}
    \put(0.7,27.5){\footnotesize $\mathbf{a}_1$}
    \put(0.7,23.5){\footnotesize $\mathbf{c}_1$}
    \put(5.3,23.4){\footnotesize $0$}
    \put(25.9,23.4){\footnotesize \textcolor{white}{$1$}}
    \put(9,24.5){\footnotesize ...}
    \put(20,24.5){\footnotesize \textcolor{white}{...}}
    
    \put(15,19){\rotatebox{90}{\footnotesize ...}}

    \put(5.3,18.4){\footnotesize $1$}
    \put(31.5,18.4){\footnotesize $\ell_N$}
    \put(0.3,14){\footnotesize $\mathbf{x}_N$}
    \put(0.3,8.5){\footnotesize $\mathbf{a}_N$}
    \put(0.3,4.5){\footnotesize $\mathbf{c}_N$}
    \put(5.3,4.4){\footnotesize $0$}
    \put(31.8,4.4){\footnotesize \textcolor{white}{$1$}}
    \put(12.3,5.5){\footnotesize ...}
    \put(23,5.5){\footnotesize \textcolor{white}{...}}
  
    \put(39,14){\footnotesize $a$}
    \put(39,6.5){\footnotesize $c$}
    \put(39,27){\footnotesize $x$}
    \put(59,22){\footnotesize $\mu$}
    \put(59,16.5){\footnotesize $\sigma$}
    \put(44,19){\footnotesize Encoder}
    \put(54.5,28){\footnotesize (reparametrization)}
    \put(67.5, 24.5){$\epsilon$}

    \put(62.5,36.5){\footnotesize $\mathcal{L}_{\text{recon}}$}
    \put(96.5,27){\footnotesize $\hat{x}$}
    \put(74.5,10){\footnotesize $a$}
    \put(74.5,3){\footnotesize $c$}
    \put(74.5,20){\footnotesize $z$}
    \put(80,19.5){\footnotesize Decoder}
    \put(61.3,7.5){\footnotesize $\mathcal{L}_{\text{reg}}$}
    \end{overpic}
    \caption{Temporally Coherent Action (TCA) model. The input to the encoder is the concatenation of the frame feature $x$, action label $a$ and coherence variable $c$. The decoder samples a latent variable with reparametrization: $z = \mu + \sigma \odot \epsilon, \epsilon \sim \mathcal{N}(\mathbf{0},\mathbf{1})$, and outputs $\hat{x}$ as the reconstruction of the original feature. }
    \label{fig:tca}
\end{figure}
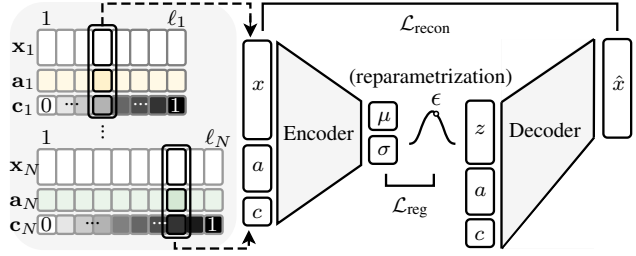

\noindent \textbf{Temporal Coherence Modeling.}
The above action model only captures the diversity of action segments but ignores any temporal coherence since the action frames are modeled independently. To that end, we introduce a coherence variable to model the temporal transition between feature frames within an action segment. The coherence is defined as the relative temporal progression of a frame within the action. For the $i$-th frame in an action segment of duration $\ell$, the coherence variable $c_i$ is defined as: 
\begin{equation}\label{eq:covar}
    c_i = (i-1)/(\ell-1), \;\text{and}\; c_i \in [0, 1].
\end{equation}
Incorporating the relative progression as the coherence condition helps the model build up knowledge of the feature continuity following the progression of actions. In this paper, we follow~\cite{becattini2020done} and assume the action progression to be linear.  
Adding upon~\cref{eq:cvae}, the training loss for our Temporally Coherent Action (TCA) model is written as:
\begin{equation}\label{eq:tca}
    \mathcal{L}_{\text{TCA}} = \underbrace{\mathbb{E}_z \log p_{\theta}(x|z,a,c)}_{\mathcal{L}_{\text{recon}}} - \underbrace{\text{D}_{\text{KL}}(q_{\phi}(z|x,a,c)||p(z))}_{\mathcal{L}_{\text{reg}}}.
\end{equation}
\cref{fig:tca} illustrates the TCA model. For each frame in a segment, the encoder $\mathcal{E}(x,a,c) = q_{\phi}(z|x,a,c)$ takes in the feature $x$, one-hot action label $a$, and its coherence variable $c$, while the decoder $\mathcal{D}(z,a,c) = p_{\theta}(x|z,a,c)$ samples a latent variable $z$ and uses the same $a,c$ for reconstruction.

\subsection{Replay Data Generation}
Our replay data generation is top-down; it first samples a sequential structure before generating the action segments.

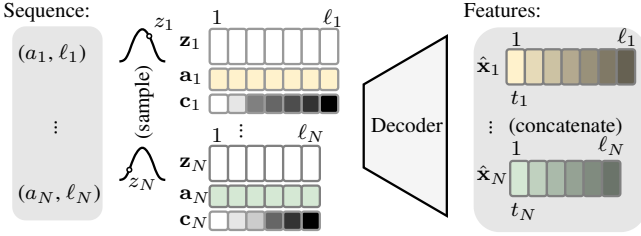
\begin{figure}
    \centering
    \begin{overpic}[width=1.02\linewidth]{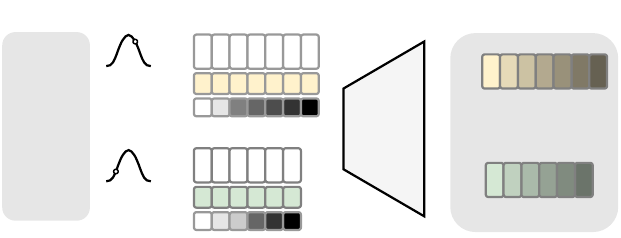}

    \put(-1,34.5){\footnotesize Sequence:}
    
    \put(1.3,28){\footnotesize ($a_1$, $\ell_1$)}
    \put(22.5,33){\footnotesize $z_1$}
    \put(26.5,30){\footnotesize $\mathbf{z}_1$}
    \put(26.5,24.5){\footnotesize $\mathbf{a}_1$}
    \put(26.5,20.5){\footnotesize $\mathbf{c}_1$}
    \put(31.5,33.5){\footnotesize $1$}
    \put(49,33.5){\footnotesize $\ell_1$}
    \put(73,26){\footnotesize $\hat{\mathbf{x}}_1$}
    \put(19.5,15.5){\rotatebox{90}{\footnotesize (sample)}}
    \put(7,16){\rotatebox{90}{\footnotesize ...}}

    \put(36,15.5){\rotatebox{90}{\footnotesize ...}}
    \put(31.5,15){\footnotesize $1$}
    \put(45,15.5){\footnotesize $\ell_N$}
    \put(75,16){\rotatebox{90}{\footnotesize ...}}
    \put(78,16.5){\footnotesize (concatenate)}
    \put(1.3,6){\footnotesize ($a_N$, $\ell_N$)}
    \put(18.5,8){\footnotesize $z_N$}
    \put(26.5,11){\footnotesize $\mathbf{z}_N$}
    \put(26.5,6){\footnotesize $\mathbf{a}_N$}
    \put(26.5,2){\footnotesize $\mathbf{c}_N$}
    \put(73,9){\footnotesize $\hat{\mathbf{x}}_N$}
    \put(71,34.5){\footnotesize Features:}
    \put(56.5,16.5){\footnotesize Decoder}
    \put(78.3,13){\footnotesize $1$}
    \put(92,13.5){\footnotesize $\ell_N$}
    \put(78.3,30){\footnotesize $1$}
    \put(95,30.5){\footnotesize $\ell_1$}
    \put(78.3,3.4){\footnotesize $t_N$}
    \put(78.3,21){\footnotesize $t_1$}

    \end{overpic}
    \caption{Replay Data Generation. In each segment, the action variable $a$ remains consistent, and a common latent variable $z\sim \mathcal{N}(\mathbf{0},\mathbf{1})$ is sampled for all frames. The resulting features $\hat{\mathbf{x}}$ by the decoder exhibit continuity due to the gradual change in the coherence variable $c$. The segments are then concatenated following their starting timestamp $t$ to form a complete video.}
    \label{fig:tca_gen}
\end{figure}

\noindent \textbf{Sequential Structure Sampling.}
As described by the segment-level interpretation provided by~\cref{eq:segment}, a procedural video can be summarized as an ordered sequence of actions, each with varying durations. Such a high-level structure helps guide the data generation process. %
The (symbolic) sequences require negligible storage, so we keep all the sequences from the training data and directly sample from the set during generation. Specifically, we establish a candidate pool to store all action sequence order and their durations $\mathbf{S}_b=\{\mathbf{s}_i\}_{i=1}^{n_{b}}, \mathbf{s}_i \in \mathcal{T}_{b}$. During the data generation stage, we employ a uniform sampling strategy on $\mathbf{S}_b$:
\begin{equation}\label{eq:sample}
    \hat{\mathbf{s}}_b \sim \text{Uniform}(\mathbf{S}_{b}).
\end{equation}

\begin{figure*}[!t]
    \centering
    \subfigure[Standard TAS]{\label{subfig:standard}
    \begin{overpic}[width=0.24\textwidth]{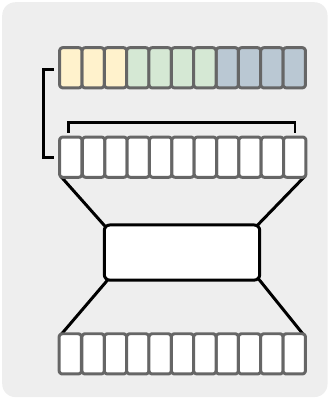} 
    \put(35,21){\footnotesize Inputs: $\mathbf{x}$}
    \put(33,50.5){\footnotesize Logits: $\mathbf{p}$}
    \put(33,93.5){\footnotesize Labels: $\mathbf{y}$}
    \put(40,72.5){\footnotesize $\mathcal{L}_{\text{sm}}$}
    \put(1.4,72.5){\footnotesize $\mathcal{L}_{\text{cls}}$}
    \put(32.5,39.5){\footnotesize  TAS Model }
    \put(40,33){\footnotesize $\mathcal{M}$}
    \end{overpic}
    }\hspace{0.5em}
    \subfigure[Incremental TAS with TCA Replay]{\label{subfig:incremental}
    \begin{overpic}[width=0.7\textwidth]{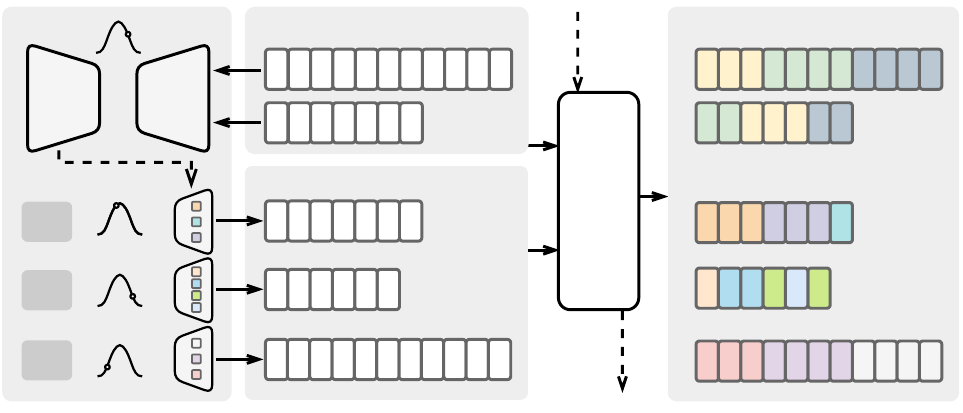}  

    \put(3.5,4.5){\footnotesize $\mathbf{S}_1$}
    \put(3.5,11.7){\footnotesize $\mathbf{S}_2$}
    \put(2.8,18.9){\footnotesize $\mathbf{S}_{b\!-\!1}$}

    \put(15.7,8){\footnotesize $\mathcal{D}_1$}
    \put(15.7,15.5){\footnotesize $\mathcal{D}_2$}
    \put(15,22.7){\footnotesize $\mathcal{D}_{b\!-\!1}$}

    \put(17,31.5){$\mathcal{E}_{b}$}
    \put(5,31.5){$\mathcal{D}_{b}$}
    \put(14,39.5){\footnotesize $\epsilon$}
    \put(2,39.5){\footnotesize TCA}
    \put(9.3,26.9){\footnotesize (cache)}

    \put(30,39.5){\footnotesize Current Task Data $\mathcal{T}_b$}
    \put(47,30){\footnotesize ...}
    \put(27,23){\footnotesize Generated Replay Data $\hat{\mathcal{T}}_{1:b}$}
    \put(47,20){\footnotesize ...}
    \put(45,13){\footnotesize ...}

    \put(63, 26.5){\footnotesize \rotatebox{270}{TAS Model}}
    \put(60, 22.5){\footnotesize \rotatebox{270}{$\mathcal{M}$}}

    \put(61.5,39.5){\footnotesize Task}
    \put(61.5,37.2){\footnotesize $b\!-\!1$}
    \put(58,4.5){\footnotesize Task}
    \put(58,2.3){\footnotesize $b\!+\!1$}

    \put(75,39.5){\footnotesize Segmentation Results}
    \put(92,20){\footnotesize ...}
    \put(90,13){\footnotesize ...}
    \put(92,30){\footnotesize ...}
    \end{overpic}}
    \caption{Learning schemes for temporal action segmentation. (a) Standard TAS and (b) Incremental TAS with TCA at task $b$. }
    \label{fig:archi}
\end{figure*}
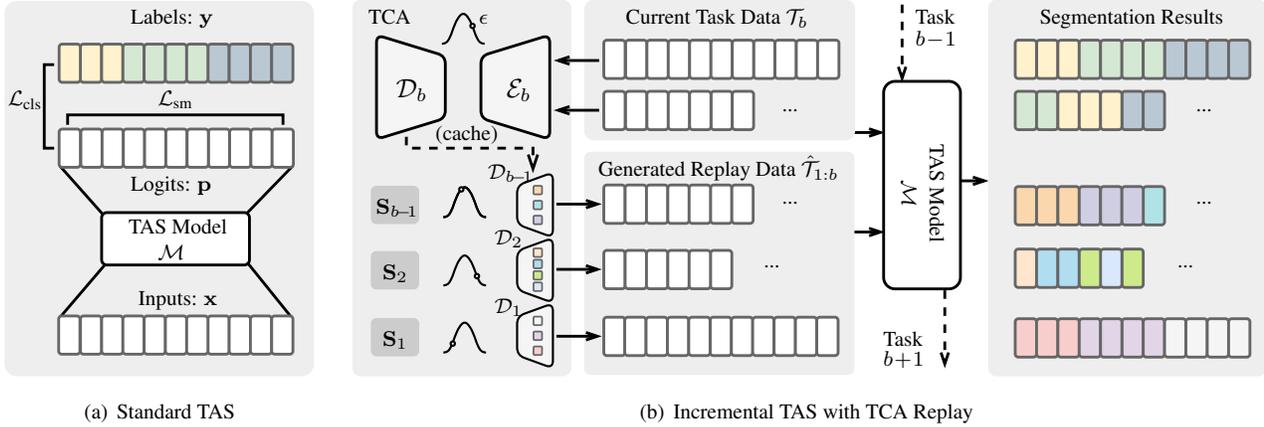

\begin{algorithm}[t]
\caption{Incremental Temporal Action Segmentation}
\textbf{Input:} Task video data $\{\mathcal{T}_1, ... ,\mathcal{T}_B \}$, replay size $M$\\
\textbf{Output:} Segmentation model $\mathcal{M}$
\begin{algorithmic}[1]
\State Train $\mathcal{M}$ with data $\mathcal{T}_1$;\Comment{\cref{eq:tas}}
\State Train $(\mathcal{E}_1, \mathcal{D}_1)$ with data $\mathcal{T}_1$;\Comment{\cref{eq:tca}}
\For{$b = 2,...,B$}\Comment{\textit{Incremental Learning}}
    \State Initialize replay data $\hat{\mathcal{T}}_{\text{1:b}} = \O$;
    \For{$b' = 1,...,b-1$}\Comment{\textit{Replay Data Generation}}
    \State Get $\hat{\mathbf{s}}_{b'}$ from $\mathbf{S}_{b'}$ for $M/(b\!-\!1)$ times; \Comment{\cref{eq:sample}}
    \State Generate $\hat{v}_{b'}$ for each $\hat{\mathbf{s}}_{b'}$ with $\mathcal{D}_{b'}$;\Comment{\cref{eq:gen}}
    \State $\hat{\mathcal{T}}_{\text{1:b}} \gets \hat{\mathcal{T}}_{\text{1:b}} \cup \{\hat{\mathbf{v}}_{b'}\}_{M/(b-1)}$;
    \EndFor
    \State Train $\mathcal{M}$ with data $\mathcal{T}_b \cup \hat{\mathcal{T}}_{\text{1:b}}$;\Comment{\cref{eq:iltas}}
    \State Train $(\mathcal{E}_b, \mathcal{D}_b)$ with data $\mathcal{T}_b$;\Comment{\cref{eq:tca}}
\EndFor
\end{algorithmic}
\label{alg:train}
\end{algorithm} 

\noindent \textbf{Action Segment Generation.} 
The sample sequential structure $\hat{\mathbf{s}} = \{(a_n, t_n, \ell_n)\}_{n=1}^{\hat{N}}$ is then utilized in the segment generation process, as illustrated in~\cref{fig:tca_gen}.  
For each sampled $(a_n, t_n, \ell_n)$, our TCA model generates the frame-wise feature for the segment using:
\begin{equation}\label{eq:segfeat}
   \hat{x}_i = p_{\theta}(x|z_n, a_n, c_i),\; \text{and}\; {i \in [1,...,\ell]}.
\end{equation}
Here, we fix $z_n$ across all frames for the given action $a_n$ and vary the coherence variable $c_i$, following~\cref{eq:covar}. This ensures temporal coherence in feature transition within the segment. The process is repeated for every segment, and these generated segment features $\hat{\mathbf{x}}_n$ are concatenated according to their timestamps $t_n$ to compose the procedural video:
\begin{equation}\label{eq:gen}
    \hat{v} = \text{concat}(\hat{\mathbf{x}}_1, ..., \hat{\mathbf{x}}_{\hat{N}}).
\end{equation}

\noindent \textbf{Discussion.} 
An intuitive interpretation of the latent and conditioning variables in~\cref{eq:segfeat} can be as follows: $z$ governs the overall scene context, $a$ regulates the action semantics, and $c$ manages the temporal progression of the action features. Since the action label $a$ is predefined for each segment, it is kept consistent. However, multiple generation options for the action segments can be achieved by manipulating the latent variable $z$ and the conditioning coherence variable $c$.   
For instance, we can generate a \textit{static} segment with a constant latent variable $z$ and a constant coherence $c$ spanning the entire segment. On the other hand, by fixing $c$, we can generate a dynamic yet \textit{random} segment with $z_i$ independently sampled for each frame. We show in \cref{subsec:ablation} that both diversity and coherence are essential in video data replay. 
We ensure temporal coherence within segments but not across action transitions because boundary frames, as indicated in~\cite{souri2022robust,ding2022leveraging}, are ambiguous and may not be conducive to learning the segmentation model. %

\subsection{Incremental Training}
We commence by training the segmentation model $\mathcal{M}$ (\cref{subfig:standard}) and our TCA model ($\mathcal{E}_1, \mathcal{D}_1$) with the initial task data $\mathcal{T}_1$. As new task data $\mathcal{T}_b$ becomes available, we create replay video data $\hat{v}$ for all previous tasks $[1\!:\!b)$ utilizing their corresponding TCA decoders $\mathcal{D}$. This process yields a total of $M$ videos denoted as $\hat{\mathcal{T}}_{1:b}$.
To train the segmentation model $\mathcal{M}$, we combine this generated data with the real data from $\mathcal{T}_b$ to serve as the training set, as shown in~\cref{subfig:incremental}. Through this, the segmentation model can revisit the previous tasks to mitigate catastrophic forgetting while learning new tasks. The incremental learning objective at task $b$ is as follows: 
\begin{equation}\label{eq:iltas}
    \mathcal{L}^b_{\text{iltas}} = \underset{(x,y)\in\mathcal{T}_b\cup \hat{\mathcal{T}}_{1:b}}{\mathcal{L}_{\text{cls}}(x,y)} + \lambda \cdot \underset{(x,y)\in\mathcal{T}_b\cup \hat{\mathcal{T}}_{1:b}}{\mathcal{L}_{\text{sm}}(x,y)},
\end{equation}
we set $\lambda=0.15$, following~\cite{farha2019ms}. Concurrently, as iterating through the above steps, we train our TCA model $(\mathcal{E}_b, \mathcal{D}_b)$ every time new video data from task $b$ arrives. The overall incremental learning procedure is summarized in~\cref{alg:train}.

\section{Experiments}\label{sec:exp}

\subsection{Datasets and Evaluation} 

\noindent \textbf{Datasets.} We adapt Breakfast~\cite{kuehne2014language} and YouTube Instructional~\cite{alayrac2016unsupervised} datasets to incremental learning. Note other TAS datasets such as GTEA~
\cite{fathi2011learning} and 50Salads~\cite{stein2013combining} are unsuitable because most of their videos share a common set of actions, making it challenging to implement a meaningful number of incremental tasks without actions overlapping. %
\textbf{Breakfast}~\cite{kuehne2014language} dataset comprises 1,712 undirected breakfast preparation videos. There are 10 activities and a total of 48 action classes; each video features 5 to 14 actions. Each activity comprises around 150 videos for training and $20\!-\!30$ for testing. 
We use the I3D~\cite{carreira2017quo} feature representations and evaluate with the standard splits.  
\textbf{YouTube Instructional (YTI)} dataset~\cite{alayrac2016unsupervised} 
features 150 instructional videos of 5 activities (30 each). There are a total of 46 actions, each activity featuring 6 to 13 actions. 
We use 80\% of the videos in each activity for training and reserve the remaining for evaluation. We use the same set of feature representations as~\cite{sener2018unsupervised,kukleva2019unsupervised}. %

\noindent \textbf{Incremental Learning.} We partition the Breakfast and YTI datasets based on activities, each activity as a separate task, and train models incrementally. We experiment with both disjoint and blurry incremental settings on Breakfast. Disjoint tasks consider shared actions different, \eg, \emph{`take plate'} in \emph{``friedegg''} is a different class from \emph{`take plate'} in \emph{``sandwich''}. The blurry setting allows overlapping and assigns common actions across tasks the same class label.

\begin{table*}[!ht]
\centering
\scalebox{0.9}{
\begin{tabular}{clcccccccccc}
\toprule
\multirow{2}{*}{\# Tasks} & \multirow{2}{*}{} & \multicolumn{5}{c}{MSTCN~\cite{farha2019ms}} & \multicolumn{5}{c}{ASFormer~\cite{yi2021asformer}} \\ \cmidrule(lr){3-7} \cmidrule(lr){8-12}
 &  & Acc & Edit & \multicolumn{3}{c}{F1 @ \{10, 25, 50\}} & Acc & Edit & \multicolumn{3}{c}{F1 @ \{10, 25, 50\}} \\ \midrule
 &&\multicolumn{10}{c}{Breakfast}\\\midrule
\multirow{4}{*}{10} & Finetune & 7.4 & 7.2 & 7.5 & 7.0 & 5.4 & 9.9 & 9.8 & 10.3 & 9.4 & 7.5 \\
 & Exemplar~\cite{alssum2023just} & 16.1 & 13.3 & 13.8 & 12.5 & 9.5 & 12.4 & 11.2 & 11.7 & 10.7 & 8.5 \\

 &  \cellcolor{gray!30}Ours & \cellcolor{gray!30}\textbf{29.4} & \cellcolor{gray!30}\textbf{25.9} & \cellcolor{gray!30}\textbf{26.3} & \cellcolor{gray!30}\textbf{23.5} & \cellcolor{gray!30}\textbf{17.7} & \cellcolor{gray!30}\textbf{34.2} & \cellcolor{gray!30}\textbf{32.4} & \cellcolor{gray!30}\textbf{33.1} & \cellcolor{gray!30}\textbf{30.1} & \cellcolor{gray!30}\textbf{23.4} \\
 & \textcolor{gray}{Original} & \textcolor{gray}{43.1} & \textcolor{gray}{41.1} & \textcolor{gray}{41.2} & \textcolor{gray}{37.6} & \textcolor{gray}{29.5} & \textcolor{gray}{48.1} & \textcolor{gray}{45.2} & \textcolor{gray}{45.9} & \textcolor{gray}{42.4} & \textcolor{gray}{34.2} \\ \midrule
\multirow{4}{*}{5} & Finetune & 15.4 & 15.8 & 16.6 & 15.8 & 12.7 & 15.7 & 16.1 & 16.9 & 15.8 & 13.2 \\
 & Exemplar~\cite{alssum2023just} & 32.5 & 28.9 & 30.8 & 28.5 & 22.9 & 29.5 & 27.5 & 28.7 & 26.7 & 22.0 \\
 & \cellcolor{gray!30}Ours & \cellcolor{gray!30}\textbf{54.5} & \cellcolor{gray!30}\textbf{49.4} & \cellcolor{gray!30}\textbf{51.1} & \cellcolor{gray!30}\textbf{46.9} & \cellcolor{gray!30}\textbf{37.7} & \cellcolor{gray!30}\textbf{57.2} & \cellcolor{gray!30}\textbf{56.8} & \cellcolor{gray!30}\textbf{58.3} & \cellcolor{gray!30}\textbf{54.0} & \cellcolor{gray!30}\textbf{43.6} \\
 & \textcolor{gray}{Original} & \textcolor{gray}{60.4} & \textcolor{gray}{59.1} & \textcolor{gray}{60.3} & \textcolor{gray}{56.1} & \textcolor{gray}{46.0} & \textcolor{gray}{65.1} & \textcolor{gray}{64.2} & \textcolor{gray}{65.6} & \textcolor{gray}{61.5} & \textcolor{gray}{51.0} \\ \midrule
&&\multicolumn{10}{c}{YouTube Instructional}\\ \midrule
\multirow{4}{*}{5}& Finetune & 13.6 & 2.8 & 3.6 & 2.7 & 0.6 & 13.9 & 11.5 & 11.1 & 9.8 & 6.3 \\
& Exemplar~\cite{alssum2023just} & \textbf{30.8} & 19.7 & 19.8 & 16.0 & 9.3 & 22.1 & 18.9 & 17.7 & 15.3 & 10.0 \\
& \cellcolor{gray!30}Ours & \cellcolor{gray!30}30.2 & \cellcolor{gray!30}\textbf{25.0} & \cellcolor{gray!30}\textbf{21.9}& \cellcolor{gray!30}\textbf{18.5} & \cellcolor{gray!30}\textbf{11.1} & \cellcolor{gray!30}\textbf{25.2} & \cellcolor{gray!30}\textbf{20.9} & \cellcolor{gray!30}\textbf{20.1} & \cellcolor{gray!30}\textbf{17.5} & \cellcolor{gray!30}\textbf{11.4} \\
&\textcolor{gray}{Original} & \textcolor{gray}{55.9} & \textcolor{gray}{39.4} & \textcolor{gray}{38.1} & \textcolor{gray}{32.2} & \textcolor{gray}{19.1} & \textcolor{gray}{59.2} & \textcolor{gray}{51.1} & \textcolor{gray}{45.4} & \textcolor{gray}{39.1} & \textcolor{gray}{25.5} \\ \bottomrule
\end{tabular}}
\caption{Performance comparison on Breakfast and YouTube Instructional. Our approach consistently surpasses the baselines, with both MSTCN and ASFormer backbones. %
}
\vspace{-1em}
\label{tab:all}
\end{table*}

\noindent \textbf{Evaluation Measures.} 
TAS is evaluated by a frame-wise accuracy (Acc), segment-wise edit score (Edit), and F1 score with varying overlap thresholds of 10\%, 25\%, and 50\%. We adopt these standard measures and apply them to the incremental setup. We denote the Acc on task $b'$ after the $b$-th task ($b'< b$) as $\text{Acc}_b^{b'}$, and use the last stage accuracy averaged over all asks as the final metric to measure the overall performance\footnote{We calculate the average performance over all tasks due to the imbalance in frame numbers across each task.}, \ie, 
\begin{equation}
    \text{Acc} = \sum_{b=1}^B \text{Acc}_B^b. 
\end{equation}
The Edit and F1 scores are similarly defined.

\subsection{Implementations}
\noindent \textbf{TCA Model.}
Our TCA model comprises an encoder and decoder; each implemented as a two-layer MLP with a 256-D latent space. Based on their dataset size, we train TCA for 2,500 epochs with Breakfast and 250 epochs for YTI. The learning rate is set to be $1e^{-3}$ and $1e^{-5}$, respectively. 

\noindent \textbf{TAS Backbones.} 
We experiment with the temporal convolutional network MSTCN~\cite{farha2019ms} and the transformer-based ASFormer~\cite{yi2021asformer}
for our backbone model $\mathcal{M}$. MSTCN is trained with a learning rate of $5e^{-4}$ for 50 epochs for each task; for ASFormer, it is $1e^{-4}$ for 30 epochs. 

\noindent \textbf{Baselines.} 
We first use a naive (\textbf{`Finetune'}) baseline that progressively finetunes with only training data in the task sequence without data replay. Additionally, we follow the boring baseline from action recognition~\cite{alssum2023just} and store a mean frame for each action segment per sequence, which we call \textbf{`Exemplar'}. Such a segment is static, as the frame-wise sample is simply replicated to inflate the segment in time and used for data replay. Finally, we consider an upper-bound (\textbf{`Original'}) data replay baseline that uses the original sequence features.

\noindent \textbf{Replay Size.} 
For all experiments, we constrain the maximum number of videos for replay to be 60\footnote{We choose the value of 60 to maintain a similar exemplar-to-training ratio (1:25) per iCIFAR-100 used in~\cite{rebuffi2017icarl}.}, which is equally divided by the number of seen tasks. Given that our sequence is sampled from seen videos, the total frame number in the generated set is bounded by the $60\times$ longest video. 
Our ablation study (see \cref{para:rs}) shows that the performance can be further improved with a larger replay size.

\begin{table*}[ht!]
\centering
\scalebox{0.9}{
\begin{tabular}{lcccccccccc}
\toprule
\multirow{2}{*}{} & \multicolumn{5}{c}{MSTCN~\cite{farha2019ms}} & \multicolumn{5}{c}{ASFormer~\cite{yi2021asformer}} \\ \cmidrule(lr){2-6} \cmidrule(lr){7-11} 
 & Acc & Edit & \multicolumn{3}{c}{F1 @ \{10, 25, 50\}} & Acc & Edit & \multicolumn{3}{c}{F1 @ \{10, 25, 50\}} \\ \cmidrule(lr){2-6} \cmidrule(lr){7-11}
Finetune & 15.8 & 29.7 & 30.0 & 26.0 & 19.1 & 15.5 & 30.2 & 30.6 & 26.4 & 19.6 \\
Exemplar~\cite{alssum2023just} & 25.0 & 34.0 & 34.9 & 30.2 & 22.3 & 24.4 & 35.6 & 37.2 & 32.9 & 24.9 \\
\rowcolor{gray!30}
Ours & \textbf{38.5} & \textbf{43.3} & \textbf{44.9} & \textbf{39.5} & \textbf{29.7} & \textbf{44.2} & \textbf{51.2} & \textbf{53.0} & \textbf{47.6} & \textbf{36.8} \\
\textcolor{gray}{Original} & \textcolor{gray}{48.5} & \textcolor{gray}{53.4} & \textcolor{gray}{55.1} & \textcolor{gray}{49.4} & \textcolor{gray}{37.9} & \textcolor{gray}{53.5} & \textcolor{gray}{57.6} & \textcolor{gray}{59.9} & \textcolor{gray}{54.5} & \textcolor{gray}{43.2} \\ \bottomrule
\end{tabular}}
\vspace{-0.5em}
\caption{Performance comparsion on Breakfast with the blurry task boundary.}
\label{tab:blurry}
\end{table*}

\subsection{Effectiveness}
We compare the performances on two datasets in~\cref{tab:all}. Finetune achieves the lowest performance as it simply concentrates on learning new concepts without revisiting previous tasks. 
In the 10-task setup on Breakfast, static segments with stored exemplar features mitigate forgetting, gaining an accuracy increase from 7.4\% to 16.1\% with MSTCN. Our approach achieves a more substantial boost, reaching 29.4\%. The improvement is consistent across both backbones. Despite this improvement, there remains a 13.1\% gap compared to using original features from previous tasks.
There are similar performance differences across the approaches in the 5-task incremental step. The 5-task incremental setup combines every two activities as a single task and has fewer training steps, improving performance across all approaches. This suggests that the challenge of learning increases with the number of tasks, leading to increased forgetting by the model. %
The performance improvement for the YTI dataset is less significant compared to Breakfast; both the Exemplar~\cite{alssum2023just} and our method achieve very similar performances. We hypothesize that this phenomenon results from the overall reduced segment diversity in the YTI dataset due to its comparatively smaller size. Nevertheless, the rise in the segmental metrics suggests that temporally evolving action segment features, as opposed to static ones, can alleviate over-segmentation. 
When comparing the backbones, ASFormer~\cite{yi2021asformer} is prone to overfit to static data compared to MSTCN~\cite{farha2019ms}. The performance on Breakfast demonstrates that ASFormer outperforms MSTCN across all approaches except in Exemplar, where static segments are employed for replay.  %

\noindent \textbf{Blurry CIL.} 
Our evaluation using the blurry setup on the Breakfast dataset is presented in~\cref{tab:blurry}. We consistently observe performance improvements, with all performance metrics surpassing the standard setup (\cref{tab:all}) by 10 points. This improvement is likely attributed to the efficient updating of action classifiers from previous tasks with the current task samples due to shared actions and blurred task boundaries.

\subsection{Ablation Study}\label{subsec:ablation}
\noindent \textbf{Feature Diversity and Temporal Coherence.} 
In our evaluation of the impact of feature diversity and temporal coherence on performance (see \cref{tab:effectiveness}), we assess three key factors: the diversity of actions at the segment level, the diversity of frames within a segment, and the temporal coherence of a segment. Both with static segments, Ours$_\text{static}$ outperforms the Exemplar by approximately 10\% in accuracy (Rows 1 \textit{vs.} 3), which showcases TCA's ability to generate diverse actions and enhance segment-level diversity. However, Ours$_\text{random}$, which maximizes feature diversity but lacks temporal constraints, performs less effectively than Ours$_\text{static}$, highlighting the detrimental impact the absence of temporal constraints can cause for the TAS task. In the final row, our approach, which coordinates feature diversity and temporal coherence, achieves the highest performance. 

\newcommand{\improv}[1]{\small\textcolor{red}{#1}}
\begin{table}[!t]
\centering
\resizebox{\linewidth}{!}{
\begin{tabular}{lcccccccc}
\toprule
 & SD & FD & TC & Acc &  Edit &  \multicolumn{3}{l}{F1 @ \{10, 25, 50\}} \\ \cmidrule(lr){2-4} \cmidrule(lr){5-9}
Exemplar & \cmark & \xmark & \xmark & 27.8 & 35.6 & 36.1 & 31.7 & 24.3 \\
Ours$_\text{random}$  & \cmark & \cmark & \xmark  & 32.9 & 38.9 & 40.0 & 35.6 & 27.2 \\
Ours$_\text{static}$   & \cmark & \xmark & \xmark & 37.9 & 42.9 & 43.8 & 38.9 & 29.0 \\
\rowcolor{gray!30}\cellcolor{white}
Ours     & \cmark & \cmark & \cmark  & \textbf{41.8} & \textbf{45.0} & \textbf{47.0} & \textbf{41.5} & \textbf{32.0} \\ \bottomrule
\end{tabular}}
\caption{Ablation study on the components in TCA.  `SD' denotes the diveristy between segments of the same action, `FD' denotes the diversity between the frames within the same segment, and `TC' denotes the temporal coherence of the segments. Optimal results are obtained when considering both diversity and temporal coherence. Gray row indicates the final setup used in this paper.}
\label{tab:effectiveness}
\end{table}

\begin{table}[!t]
\centering
\begin{tabular}{lccccc}
\toprule
$M$ & Acc &Edit & \multicolumn{3}{c}{F1 @ \{10, 25, 50\}} \\ \cmidrule(lr){1-1} \cmidrule(lr){2-6}
30 &  34.0 &  39.6 &  41.0 &  34.8 & 24.7 \\ 
\rowcolor{gray!30}
60 &  35.4 &  41.2 &  42.3 &  36.0 & 25.6 \\ 
90 &  36.2 &  \textbf{42.3} &  43.9 &  \textbf{37.3} & \textbf{26.8} \\ 
120 &  \textbf{38.0} &  \textbf{42.3} &  \textbf{44.0} &  37.1 & 26.2 \\ \bottomrule
\end{tabular}
\caption{Different replay sizes on Breakfast. Revisiting more replay videos can help the model mitigate forgetting.}
\label{tab:mem}
\end{table}

\noindent\textbf{Replay Size.}\label{para:rs}
We examine four different replay sizes: $M=30,60,90,120$. As depicted in~\cref{tab:mem}, performance is observed to be correlated with the replay size $M$ as anticipated. Employing only 30 video sequences as replay data results in a marginal decline (34.0\%) compared to 60 (35.4\%). An increase in the replay size $M$ corresponds to incremental improvements in both frame-wise and segmental metrics, and the best performance is achieved with the replay size being set to 120.

\noindent \textbf{TCA Training Data.}
The results of training TCA with varying proportions of task data  $\mathcal{T}$ are presented in~\cref{tab:vaeratio}. A consistent trend indicates that increased access to task data during TCA training brings greater overall performance improvements and less forgetting. Training TCA with the entire task data yields the overall best performance, with a significant 6\% decrease in performance observed when only a quarter of the data is utilized. Notably, our approach outperforms the Exemplar significantly, even when trained with only 25\% of task data. 
\begin{table}[tb]
\centering
\begin{tabular}{lcccccc}
\toprule
&$\mathcal{T}$(\%)  & Acc & Edit & \multicolumn{3}{c}{F1 @ \{10, 25, 50\}} \\ \cmidrule(lr){2-2} \cmidrule(lr){3-7}
Exemplar& - & 22.6 & 34.8 & 36.0 & 32.4 & 25.2\\ \cmidrule(lr){2-2} \cmidrule(lr){3-7}
\multirow{4}{*}{Ours}&25 & {41.7} & {43.2} & {46.1} & {40.9} & 31.5 \\ 
&50 & {42.1} & {43.3} & {45.1} & {40.5} & 31.5 \\ 
&75 & {45.3} & {45.9} & {47.8} & \textbf{43.7} & \textbf{34.7} \\ 
\rowcolor{gray!30}\cellcolor{white} & 100 & \textbf{47.4} & \textbf{46.9} & \textbf{48.2} & {42.8} & 33.4 \\ \bottomrule
\end{tabular}
\caption{Different ratios for TCA training on Breakfast. The final performance is correlated with the quantity of data used in TCA training. A higher volume of training data leads to improved final perfromance.}
\label{tab:vaeratio}
\end{table}

\begin{figure}
    \centering
    \begin{overpic}[width=0.95\linewidth]{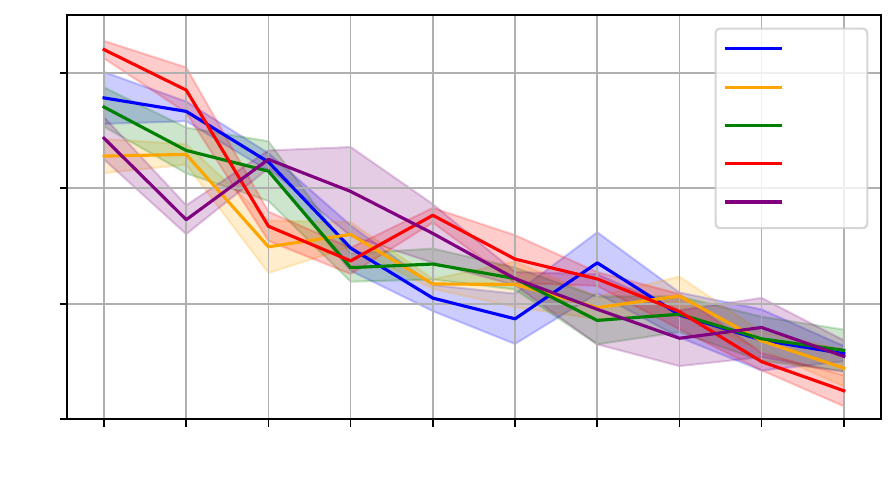}
    \put(2.5,7){\footnotesize 20}
    \put(2.5,19){\footnotesize 40}
    \put(2.5,32){\footnotesize 60}
    \put(2.5,45){\footnotesize 80}
    \put(10.5,3){\footnotesize 1}
    \put(20,3){\footnotesize 2}
    \put(29,3){\footnotesize 3}
    \put(38,3){\footnotesize 4}
    \put(47.5,3){\footnotesize 5}
    \put(57,3){\footnotesize 6}
    \put(66,3){\footnotesize 7}
    \put(75,3){\footnotesize 8}
    \put(84,3){\footnotesize 9}
    \put(93,3){\footnotesize 10}
    \put(45,-1.8){\footnotesize Incremental Step}
    \put(-2,27){\footnotesize \rotatebox{90}{Acc}}
    \put(88,47.5){\footnotesize 42}
    \put(88,43){\footnotesize 123}
    \put(88,38.7){\footnotesize 1000}
    \put(88,34.5){\footnotesize 1993}
    \put(88,30.5){\footnotesize 2023}
    \end{overpic}
    \vspace{0.05em}
    \caption{Performance following each incremental step with the varied task sequence. The mean and standard deviation across four splits are represented by solid lines and shaded areas, respectively. Each unique task sequence, determined by a random seed, is distinguished by a different color.}
    \label{fig:order_b10}
\end{figure}

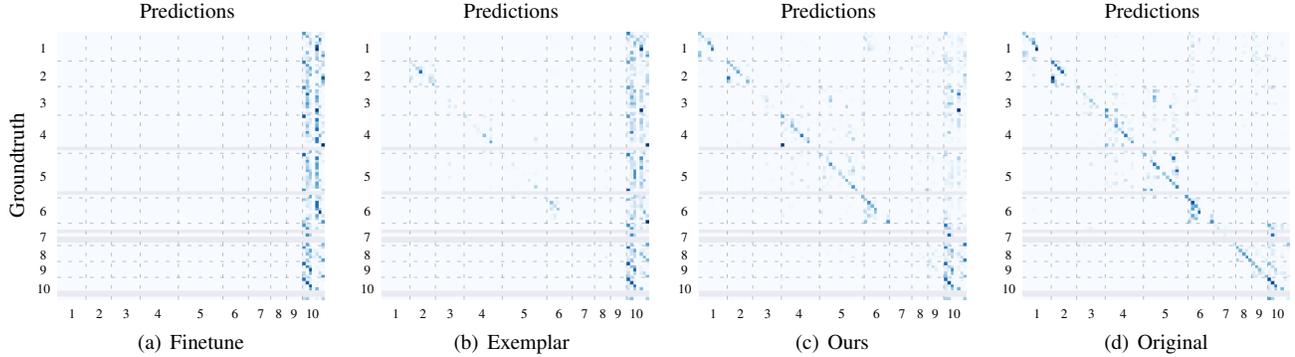
\begin{figure*}[!t]
\centering
\hspace{0.75em}
\subfigure[Finetune]{\label{subfig:finetune}
\begin{overpic}[width=0.23\textwidth]{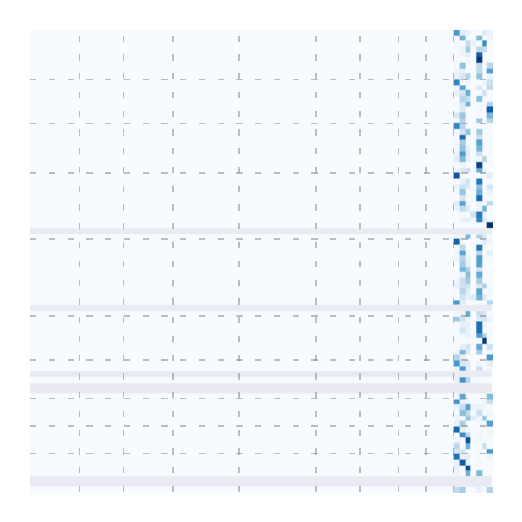}
\put(-10,33){\rotatebox{90}{\footnotesize Groundtruth}}
\put(33,99){\rotatebox{0}{\footnotesize Predictions}}
\put(9.5,-0.5){\tiny 1}
\put(18.5,-0.5){\tiny 2}
\put(27.5,-0.5){\tiny 3}
\put(38,-0.5){\tiny 4}
\put(52,-0.5){\tiny 5}
\put(63.5,-0.5){\tiny 6}
\put(72,-0.5){\tiny 7}
\put(78,-0.5){\tiny 8}
\put(83,-0.5){\tiny 9}
\put(88,-0.5){\tiny 10}
\put(0,87.5){\tiny 1}
\put(0,78.5){\tiny 2}
\put(0,69.5){\tiny 3}
\put(0,59){\tiny 4}
\put(0,45){\tiny 5}
\put(0,33.5){\tiny 6}
\put(0,25){\tiny 7}
\put(0,19){\tiny 8}
\put(0,14){\tiny 9}
\put(-1,8){\tiny 10}
\end{overpic}
}\hfill
\subfigure[Exemplar]{\label{subfig:exemplar}
\begin{overpic}[width=0.23\textwidth]{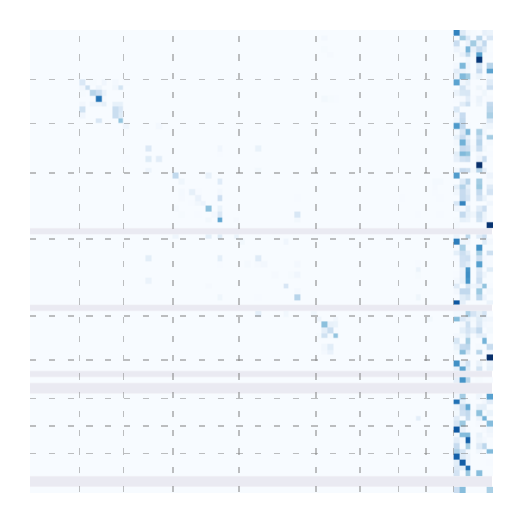}
\put(33,99){\rotatebox{0}{\footnotesize Predictions}}
\put(9.5,-0.5){\tiny 1}
\put(18.5,-0.5){\tiny 2}
\put(27.5,-0.5){\tiny 3}
\put(38,-0.5){\tiny 4}
\put(52,-0.5){\tiny 5}
\put(63.5,-0.5){\tiny 6}
\put(72,-0.5){\tiny 7}
\put(78,-0.5){\tiny 8}
\put(83,-0.5){\tiny 9}
\put(88,-0.5){\tiny 10}
\put(0,87.5){\tiny 1}
\put(0,78.5){\tiny 2}
\put(0,69.5){\tiny 3}
\put(0,59){\tiny 4}
\put(0,45){\tiny 5}
\put(0,33.5){\tiny 6}
\put(0,25){\tiny 7}
\put(0,19){\tiny 8}
\put(0,14){\tiny 9}
\put(-1,8){\tiny 10}
\end{overpic}}\hfill
\subfigure[Ours]{\label{subfig:ours}
\begin{overpic}[width=0.23\textwidth]{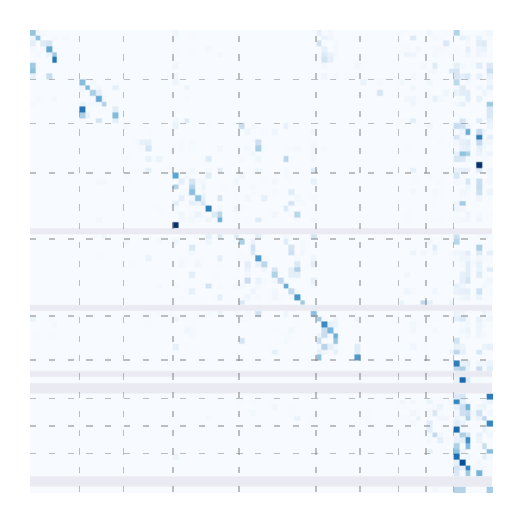}
\put(33,99){\rotatebox{0}{\footnotesize Predictions}}
\put(9.5,-0.5){\tiny 1}
\put(18.5,-0.5){\tiny 2}
\put(27.5,-0.5){\tiny 3}
\put(38,-0.5){\tiny 4}
\put(52,-0.5){\tiny 5}
\put(63.5,-0.5){\tiny 6}
\put(72,-0.5){\tiny 7}
\put(78,-0.5){\tiny 8}
\put(83,-0.5){\tiny 9}
\put(88,-0.5){\tiny 10}
\put(0,87.5){\tiny 1}
\put(0,78.5){\tiny 2}
\put(0,69.5){\tiny 3}
\put(0,59){\tiny 4}
\put(0,45){\tiny 5}
\put(0,33.5){\tiny 6}
\put(0,25){\tiny 7}
\put(0,19){\tiny 8}
\put(0,14){\tiny 9}
\put(-1,8){\tiny 10}
\end{overpic}
}\hfill
\subfigure[Original]{\label{subfig:gt}
\begin{overpic}[width=0.23\textwidth]{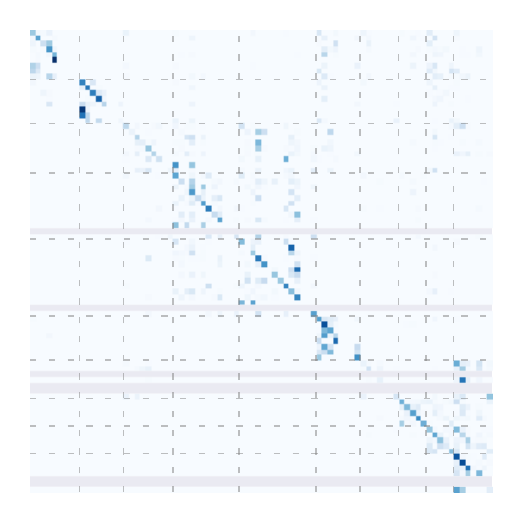}
\put(33,99){\rotatebox{0}{\footnotesize Predictions}}
\put(9.5,-0.5){\tiny 1}
\put(18.5,-0.5){\tiny 2}
\put(27.5,-0.5){\tiny 3}
\put(38,-0.5){\tiny 4}
\put(52,-0.5){\tiny 5}
\put(63.5,-0.5){\tiny 6}
\put(72,-0.5){\tiny 7}
\put(78,-0.5){\tiny 8}
\put(83,-0.5){\tiny 9}
\put(88,-0.5){\tiny 10}
\put(0,87.5){\tiny 1}
\put(0,78.5){\tiny 2}
\put(0,69.5){\tiny 3}
\put(0,59){\tiny 4}
\put(0,45){\tiny 5}
\put(0,33.5){\tiny 6}
\put(0,25){\tiny 7}
\put(0,19){\tiny 8}
\put(0,14){\tiny 9}
\put(-1,8){\tiny 10}
\end{overpic}
}
\vspace{-0.5em}
\caption{Comparsion of confusion matrices of different approaches on Breakfast in a 10-task setup. The task sequence is given as follows: 1 - ``\textit{sandwich}'',  2 - ``\textit{juice}'', 3 - ``\textit{friedegg}'', 4 - ``\textit{scrambledegg}'', 5 - ``\textit{pancake}'', 6 - ``\textit{salat}'', 7  - ``\textit{tea}'', 8 - ``\textit{milk}'', 9 - ``\textit{cereal}'', 10 - ``\textit{coffee}''. The dashed lines indicate task boundaries, while gray rows denote the lack of instances for that action in the test data. Our approach (c) shows the closest resemblance to data replay using original features (d). }
\label{fig:cm}
\end{figure*}

\noindent \textbf{Task Sequence.}
We compare five distinct task sequence arrangements within a 10-task incremental setup on Breakfast and report the results in~\cref{fig:order_b10}. 
Although there is a consistent decrease in accuracy scores, the ultimate performance is subject to variation depending on the seed used for task sequence determination, showing differences of up to 7\% points. This suggests that the forgetting in incremnetal learning is related to the order of these learning tasks. Moreover, the disparity could be exacerbated by the uneven distribution of frames across tasks, which is commonly observed in procedural videos.

\begin{figure}
    \centering
    \begin{overpic}[width=0.8\linewidth]{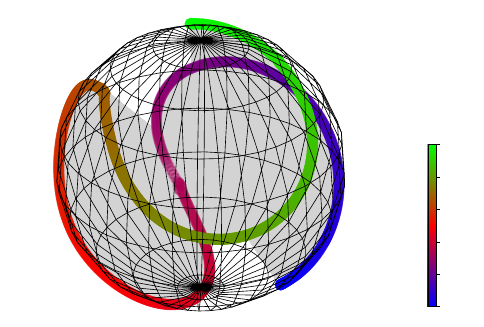}
        \put(93,3){\footnotesize 0.0}
        \put(93,9.5){\footnotesize 0.2}
        \put(93,16){\footnotesize 0.4}
        \put(93,22.5){\footnotesize 0.6}
        \put(93,29){\footnotesize 0.8}
        \put(92.9,35.5){\footnotesize 1.0}
        \put(80,5.5){ $c\!:$}
    \end{overpic}
    \caption{T-SNE visualization of a generated action segment given the action `\textit{pour milk}' from activity ``\textit{cereal}''. Every point in the sphere corresponds to a frame feature, and the continuity of these feature points demonstrates temporal coherence. }
    \vspace{-1em}
    \label{fig:tc}
\end{figure}
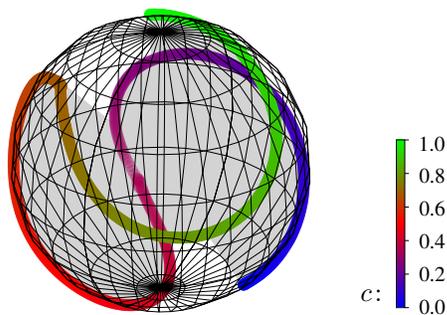

\subsection{Visualization}
\noindent \textbf{Confusion Matrix.} In~\cref{fig:cm}, we plot the confusion matrices generated by different approaches. Our method demonstrates superior performance compared to Finetune and Exemplar, particularly in the case of longer activities such as ``\textit{sandwich}'', ``\textit{scrambledegg}'', ``\textit{pancake}'' and ``\textit{salat}''.  
Furthermore, similar to Original, our approach exhibits increased confusion between semantically similar activities, such as ``\textit{scrambledegg}'' and ``\textit{pancake}'', both involving cooking with a pan.

\noindent \textbf{Temporal Coherence.} 
To assess the temporal coherence, We employ T-SNE~\cite{van2008visualizing} to visualize an exemplary action segment generated by our TCA model. Specifically, we employ the TCA model trained on ``\textit{cereal}'' data to generate an action segment of `\textit{pour milk}'. This segment comprises 1,000 frames, each sharing a common action label $a$ and latent variable $z$, with an evenly strided $c$. As shown in~\cref{fig:tc}, the features of the action segment exhibit continuous properties in the feature space, corresponding to the gradual increase in the coherence variable $c$. 

\subsection{Limitations}\label{subsec:limitation}
\vspace{0.3em}
This work presupposes the presence of dense labels for all videos within each incremental task, and our TCA model relies on such dense labels for effective learning. However, such an annotation process can be expensive in a real-world scenario. Additionally, while we can demonstrate the temporal coherence of the features generated by our TCA model, the qualitative evaluation of the features remains challenging because their existence in a feature space makes them difficult to decode into a human-readable format.

\section{Conclusion}\label{sec:conclusion}
This paper proposes a temporally coherent action model for incremental procedural video understanding from the perspective of video data replay. Departing from the conventional frame-exemplar data replay prevalent in action recognition, our approach adopts a generative model. In addition, we facilitate the modeling of temporal coherence in the actions by introducing a conditioning variable to the generative model. Our design guarantees both feature diversity and temporal coherence in the replay data, resulting in a significant performance improvement on two action segmentation benchmarks compared to the baselines. 

\section*{Acknowledgement}
This research is supported by the National Research Foundation, Singapore under its NRF Fellowship for AI (NRF-NRFFAI1-2019-0001). Any opinions, findings and conclusions or recommendations expressed in this material are those of the author(s) and do not reflect the views of National Research Foundation, Singapore.
\clearpage
{\small
\bibliographystyle{ieee_fullname}
\bibliography{egbib}
}

\end{document}


\title{Coherent Temporal Synthesis for Incremental Action Segmentation\\Supplementary Material}  %

\makeatletter
\g@addto@macro\@maketitle{
\vspace{-5em}
\begin{center}
    {\large Guodong Ding, Hans Golong and Angela Yao}\\
    \vspace{0.3em}
    {\large National University of Singapore}\\
    \vspace{0.2em}
    {\tt\small \{dinggd, hgolong, ayao\}@comp.nus.edu.sg}
\end{center}
\input{tex/tab_var}
\input{tex/fig_viz}
}
\makeatother
\maketitle

\input{tex/tab_cls}

\noindent\textbf{Multiple Runs.} We adopt the approach of prior incremental learning studies by initializing the task sequence with multiple random seeds. We utilized five specific random seeds throughout our experiments: $\{42, 123, 1000, 1993, 2023\}$. The performance results presented in the Main Paper reflect the average, while we provide the variations across runs in \cref{tab:var}.

\noindent\textbf{Total Classes.} In \cref{tab:cls}, we list the total number of action classes present in our experiments. Breakfast comprises a total of 84 actions without permitting overlaps, whereas there are 48 actions when allowing for overlapping actions. YouTube Instructionals encompasses 50 actions that do not overlap.

\noindent\textbf{Segment Visualization.} We present additional generated segments in~\cref{fig:viz}. Specifically, \cref{subfig:pmc,subfig:pmc2} depict the action of `\textit{pour milk}' generated by our TCA model for ``\textit{cereal}'' using two different latent variables $z$, highlighting the diversity in trajectories. \cref{subfig:pmp} illustrates the same `\textit{pour milk}' action in a distinct activity ``\textit{pancake}''. Furthermore, \cref{subfig:tkj} shows a segment of `\textit{take knife}' in ``\textit{juice}''. All these visualized segments indicate the temporal coherence between the frame features.

\input{tex/tab_dim}

\input{tex/tab_epoch}

\noindent\textbf{TCA Latent Space Sizes.} We varied the latent space sizes in our TCA model and presented the outcomes in~\cref{tab:dim}. The performance in incremental learning appears quite consistent across various latent space sizes. Notably, a larger dimension (512) only marginally improves performance (by less than 1\%) compared to the smaller size (128).

\noindent\textbf{TCA Training Epochs.} \cref{tab:epoch} presents TCA's performance across various training epochs. Increasing the training epoch to 5,000 yields a slight performance improvement. However, as the epoch extends to 7,500, a 3\% decline in Acc and 1\% in segmental metrics occurs. %

{\small
\bibliographystyle{ieee_fullname}
\bibliography{egbib}
}